\pdfoutput=1
\documentclass[11pt]{article}

\usepackage[preprint]{acl}

\usepackage{times}
\usepackage{latexsym}

\usepackage[T1]{fontenc}
\usepackage[utf8]{inputenc}

\usepackage{microtype}

\usepackage{inconsolata}

\usepackage{graphicx}
\usepackage{hyperref}       % hyperlinks
\usepackage{url}            % simple URL typesetting
\usepackage{amssymb}
\usepackage{booktabs}       % professional-quality tables
\usepackage{amsfonts}       % blackboard math symbols
\usepackage{nicefrac}       % compact symbols for 1/2, etc.
\usepackage{microtype}      % microtypography
\usepackage{xcolor}         % colors
\usepackage{graphicx}
\usepackage{caption}   % Needed to add subcaption
\usepackage{subcaption} % Required for subfigure and subtable
\usepackage{amsfonts}
\usepackage{microtype}
\usepackage{color}
\usepackage{multirow}
\usepackage{multicol}
\usepackage{marvosym}
\usepackage{enumitem}
\usepackage{utfsym}
\usepackage{xspace}
\usepackage{pifont}
\usepackage[T1]{fontenc}
\usepackage{times}
\usepackage{epsfig}
\usepackage{amsmath}
\usepackage{bm}
\usepackage{wrapfig}
\definecolor{tabhighlight}{HTML}{e5e5e5}
\definecolor{tabhighlight2}{HTML}{e8e8e8}
\definecolor{citecolor}{HTML}{0071BC}
\definecolor{linkcolor}{HTML}{ED1C24}

\definecolor{ForestGreen}{RGB}{34,139,34}
\definecolor{deemph}{gray}{0.6}

\usepackage{sidecap}

\usepackage{float}
\definecolor{purple}{RGB}{230, 227, 254}
\definecolor{lightgreen}{RGB}{238, 252, 241}
\definecolor{lightred}{RGB}{231, 187, 187}
\definecolor{darkred}{RGB}{198, 129, 129}

\definecolor{tabhighlight}{HTML}{e5e5e5}
\definecolor{tabhighlight2}{HTML}{e8e8e8}
\definecolor{Bittersweet}{RGB}{238, 44, 44}
\definecolor{tabhighlight}{HTML}{e5e5e5}
\definecolor{citecolor}{HTML}{0071bc}
\definecolor{trainglePurple}{HTML}{D383E0}

\usepackage{geometry}
\usepackage{caption}
\usepackage{colortbl}

\title{VANE-Bench: Video Anomaly Evaluation Benchmark \\ for Conversational LMMs}

\author{
 \textbf{Hanan Gani\textsuperscript{*1}} \quad
 \textbf{Rohit Bharadwaj\textsuperscript{*2}} \quad
  \textbf{Muzammal Naseer\textsuperscript{3}} \\
 \textbf{Fahad Shahbaz Khan\textsuperscript{1,4}} \quad
 \textbf{Salman Khan\textsuperscript{1,5}}  \\ \\
 \textsuperscript{1}Mohamed Bin Zayed University of Artificial Intelligence,
 \textsuperscript{2}University of Edinburgh,
 \\
 \textsuperscript{3}Department of Computer Science, Khalifa University,
 \\
 \textsuperscript{4}Linköping University,
 \textsuperscript{5}Australian National University
\\
 \small{
   \textbf{Correspondence:} \href{mailto:hanan.ghani@mbzuai.ac.ae}{hanan.ghani@mbzuai.ac.ae}, \href{mailto:rohit.bharadwaj@ed.ac.uk}{rohit.bharadwaj@ed.ac.uk}
 }
}

\begin{document}
\maketitle
\def\thefootnote{*}\footnotetext{Equal contribution}
\begin{abstract}
The recent advancements in Large Language Models (LLMs) have greatly influenced the development of Large Multi-modal Video Models (Video-LMMs), significantly enhancing our ability to interpret and analyze video data. Despite their impressive capabilities, current Video-LMMs have not been evaluated for anomaly detection tasks, which is critical to their deployment in practical scenarios e.g., towards identifying deepfakes, manipulated video content, traffic accidents and crimes. In this paper, we introduce VANE-Bench, a benchmark designed to assess the proficiency of Video-LMMs in detecting and localizing anomalies and inconsistencies in videos. Our dataset comprises an array of videos synthetically generated using existing state-of-the-art text-to-video generation models, encompassing a variety of subtle anomalies and inconsistencies grouped into five categories: unnatural transformations, unnatural appearance, pass-through, disappearance and sudden appearance. Additionally, our benchmark features real-world samples from existing anomaly detection datasets, focusing on crime-related irregularities, atypical pedestrian behavior, and unusual events. The task is structured as a visual question-answering challenge to gauge the models' ability to accurately detect and localize the anomalies within the videos. We evaluate nine existing Video-LMMs, both open and closed sources, on this benchmarking task and find that most of the models encounter difficulties in effectively identifying the subtle anomalies. In conclusion, our research offers significant insights into the current capabilities of Video-LMMs in the realm of anomaly detection, highlighting the importance of our work in evaluating and improving these models for real-world applications. Our code and data is publicly available at \url{https://github.com/rohit901/VANE-Bench}.
\end{abstract}

\begin{figure*}[ht]
    \centering
    \includegraphics[width=0.8\textwidth]{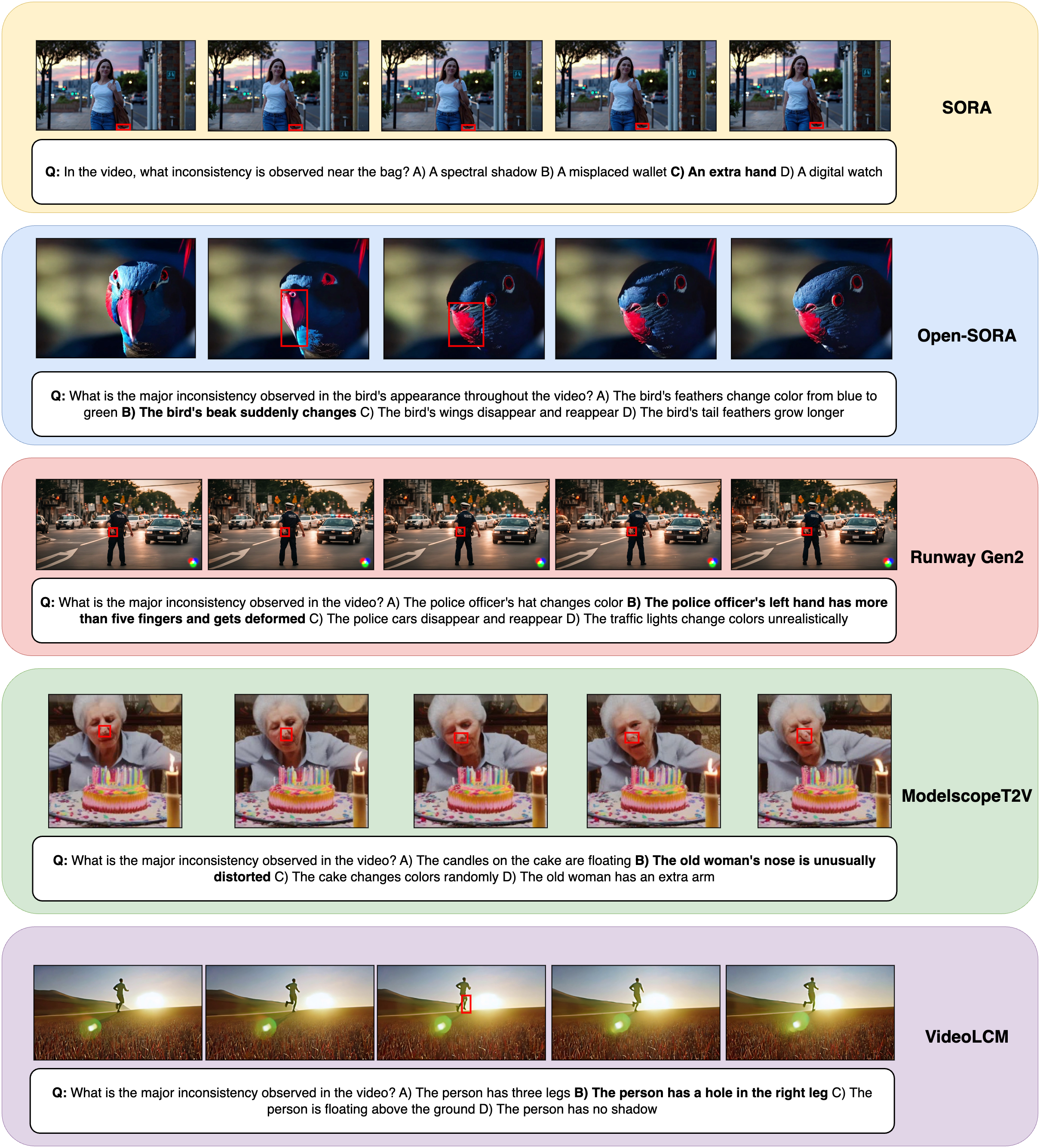}
    \caption{Samples showing the AI-Generated video category of VANE-Bench. We collect these synthetic videos from SORA~\cite{sora}, Open-Sora~\cite{open-sora}, Runway Gen2~\cite{gen2-runway}, ModelScopeT2V~\cite{wang2023modelscope}, and VideoLCM~\cite{wang2023videolcm}. The correct option in each question is highlighted in bold. Note that many of these anomalies are extremely subtle and difficult for humans to detect since the changes happen in rapid succession, with the entire video played in under a second. Anomalies are identified with red bounding boxes for clarity. Note that our actual dataset does not contain bounding box overlays.
    }
    \label{fig:intro-fig}
\end{figure*}

\section{Introduction}
Large Language Models (LLMs) like ChatGPT have ushered in a new era of real-world AI applications in varied and diverse sectors like manufacturing, legal services, space exploration, transportation, retail, healthcare, education, and technology~\cite{abdullah2022chatgpt, Marr2023}. Further, the current trend in the development of these LLMs has been to introduce multi-modal capabilities like vision and audio to these models along with text~\cite{geminiteam2024gemini, gpt4-o}. This motivates us to ask the question whether the current Large Multi-modal Models (LMMs) are capable and accurate in tackling the problem statement of Video Anomaly Detection (VAD) which has immense practical applications in factories, autonomous driving, crime warning, and traffic management~\cite{vad-survey}.

Further, we have recently observed superior visual quality of various AI-generated videos due to the improvements in the underlying algorithms, which are based on diffusion models, and transformers~\cite{sora, open-sora, peebles2023scalable}. The current state-of-the-art (SOTA) AI text-to-video model is SORA from OpenAI~\cite{sora}. The videos produced by SORA are of extremely high fidelity, which makes them nearly indistinguishable from real-life footage. Thus, SORA brings new challenges in tackling misinformation, identifying deepfakes, and distinguishing real from fake videos, especially during crucial events like democratic elections. Therefore, developing automated solutions to identify AI-generated videos has become the need of the hour.

\begin{figure*}[!t]
  \centering
  \includegraphics[scale=0.4]{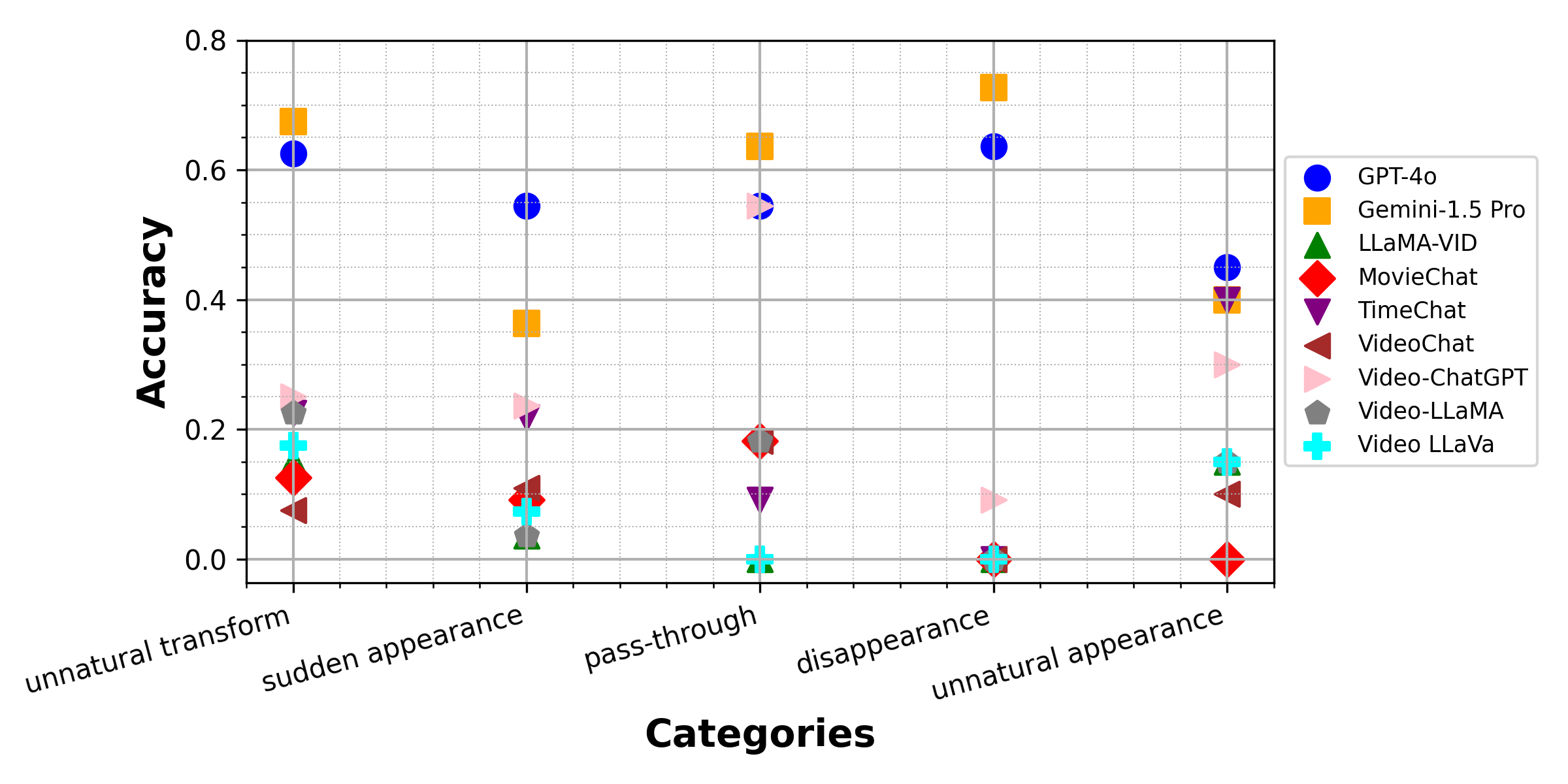}
  \includegraphics[scale=0.4]{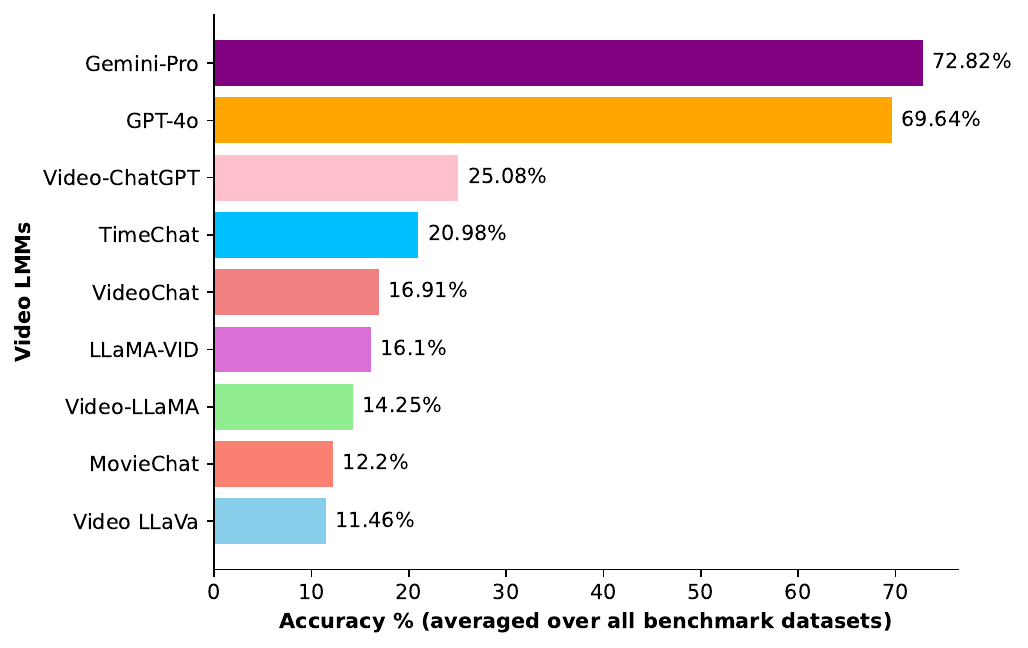}
  \caption{\textbf{Left:} Performance of Video-LMMs on five anomaly categories of SORA dataset. \textbf{Right:} Overall performance of Video-LMMs averaged across all the benchmark datasets, including AI-generated and real-world anomaly datasets.
  }
  \label{fig:intro-accuracy-all-datasets}
  \vspace{-0.3cm}
\end{figure*}

Motivated by the above-mentioned points, we propose a novel and challenging benchmark, \textbf{VANE-Bench: Video ANomaly Evaluation Benchmark}, to evaluate various closed-source and open-source Video-LMMs on their ability to detect anomalies in the videos. Our VANE-Bench consists of both real-world video anomalies from diverse surveillance footage capturing unusual pedestrian behaviour, criminal activities, and unusual events, as well as subtle and challenging anomalies and inconsistencies present in various AI-generated videos (See Fig~\ref{fig:intro-fig}). These AI-generated videos, especially from SOTA models like SORA, have subtle and hard to detect anomalies, which makes this a challenging task even for many humans. However, automatically detecting and identifying the anomalies in these synthetic video clips serves as an important step towards identifying AI-generated videos in the wild. We reformulate the problem statement of VAD into a visual question-answering (VQA) task to facilitate easier evaluation of LMMs. However, despite evaluating over nine recent Video-LMMs on VANE-Bench, we find that most current LMMs still struggle on this benchmark (see Fig.\ref{fig:intro-accuracy-all-datasets}), making VANE-Bench a challenging and a useful benchmark for tracking the progress of Video-LMMs for the foreseeable future.

Our contributions can be summarized as follows:
\begin{enumerate}
    \item We present VANE-Bench: Video ANomaly Evaluation Benchmark, consisting of 325 video clips, and 559 challenging question-answer pairs from both real-world video surveillance, and AI-generated videos.
    \item We perform detailed evaluation of over nine state-of-the-art closed-source and open-source Video-LMMs on VANE-Bench, and show that most models exhibit poor performance, highlighting the challenging nature of our proposed benchmark.
    \item We conduct detailed result analysis, and also perform human evaluation on VANE-Bench to set a reasonable benchmark target. 
    \item We will open-source our code, and describe the data construction process of VANE-Bench along with making our data publicly available.
\end{enumerate}

We hope that VANE-Bench serves as a strong benchmark to improve the performance and capabilities of Video-LMMs on anomaly detection.

\section{Related Work}
\textbf{Video-LMMs:} LMMs integrate linguistic and visual data to process videos, leveraging LLMs like Llama~\cite{llama3} and connecting them with modality-specific encoders via interfaces like Q-former~\cite{zhang2024mmllms, dai2023instructblip, yin2024survey}. Notable open-source Video-LMMs include VideoChat~\cite{li2023videochat}, which uses a chat-centric system, and VideoChatGPT~\cite{maaz2023video}, which combines a visual encoder with an LLM for detailed video conversations. Video-LLaMA~\cite{zhang2023video} integrates audio and visual signals using Q-formers, while LLaMA-VID~\cite{li2023llama} represents each frame with context and content tokens for efficient processing. Despite these advancements, our work shows current LMMs perform poorly on VANE-Bench, highlighting the need for stronger models in anomaly detection.

\noindent
\textbf{Video-LMMs Benchmarking:} Benchmarks like SEED-Bench~\cite{li2023seedbench} and MV-Bench~\cite{li2024mvbench} assess general comprehension through multiple-choice questions but lack focus on anomaly detection in AI-generated videos. CVRR-ES~\cite{Khattak2024cvrres} evaluates real-world scenarios with open-ended questions but doesn't address AI-generated inconsistencies. VANE-Bench specifically evaluates VAD in both real-world and AI-generated videos, providing a targeted benchmark for this task. While Perception Test~\cite{pătrăucean2023perception} focuses on lower-level perception in real-world videos, VANE-Bench targets subtle anomalies in AI-generated content, making it essential for assessing Video-LMM robustness.

\noindent
\textbf{Video Anomaly Detection:} Traditional VAD methods typically rely on hand-crafted features and statistical models to identify deviations from normality. CUVA~\cite{du2024uncovering} is a comprehensive benchmark that focuses on the causation of video anomalies. A survey on generalized VAD~\cite{liu2024generalized} categorizes various methodologies and highlights benchmark limitations. These methods often fail with complex AI-generated videos. VANE-Bench addresses this by focusing on VAD in such videos, complementing existing benchmarks and targeting subtle inconsistencies in high-fidelity AI-generated content.

\section{Dataset \& Benchmark}
\label{sec:datasets-and-benchmark}
Recent advancements in multi-modal Large Language Models (LLMs) have enabled these models to process text, image, and video data, presenting new opportunities and challenges in Video Anomaly Detection (VAD)~\cite{vad-survey}. Motivated by this progress, we aim to benchmark the capabilities of these multi-modal models (LMMs) on VAD.

To address VAD, we propose \textbf{VANE-Bench: Video ANomaly Evaluation Benchmark for Conversational LMMs}, comprising 325 video clips and 559 challenging ground-truth question-answer (QA) pairs. We have adapted the VAD problem into a Multiple-Choice Video Question Answering (MC-Video QA)~\cite{tapaswi2016movieqa, lei2019tvqa, yu2019activitynetqa} task to facilitate the evaluation of LMMs, allowing for a more granular assessment of their video content understanding.

We evaluate the latest closed-source and open-source LMMs on VANE-Bench. Sec.~\ref{sec:data-overview} provides an overview of VANE-Bench, Sec.~\ref{sec:data-category} describes the dataset categories, and Sec.~\ref{sec:data-collect} outlines our data collection methodology.

\subsection{Overview}
\label{sec:data-overview}

VANE-Bench consists of 325 video clips spanning real-world and synthetic video anomalies. We adapted standard VAD surveillance datasets such as CUHK Avenue~\cite{cuhk-avenue}, UCF-Crime~\cite{ucf-crime}, and UCSD Pedestrian~\cite{ucsd-ped} to our MC-Video QA problem. Additionally, we included 197 video clips from various open-source and state-of-the-art closed-source text-to-video diffusion models~\cite{sora, open-sora, gen2-runway, wang2023modelscope, wang2023videolcm}.

The diverse data backgrounds and varied difficulty levels in VANE-Bench make it ideal for evaluating the reasoning and understanding capabilities of video LMMs. Benchmarking these models on a range of real-world and synthetic anomalies helps us understand their strengths and limitations, guiding future multi-modal AI research.

Overall, VANE-Bench aims to push the boundaries of what LMMs can achieve in video anomaly detection, providing a rigorous standard for evaluating their performance on this challenging task.

\subsection{Categories}
\label{sec:data-category}

\begin{figure*}[tbp]
    \centering
    
    \begin{minipage}[b]{0.31\textwidth}
        \centering
        \includegraphics[width=\textwidth]{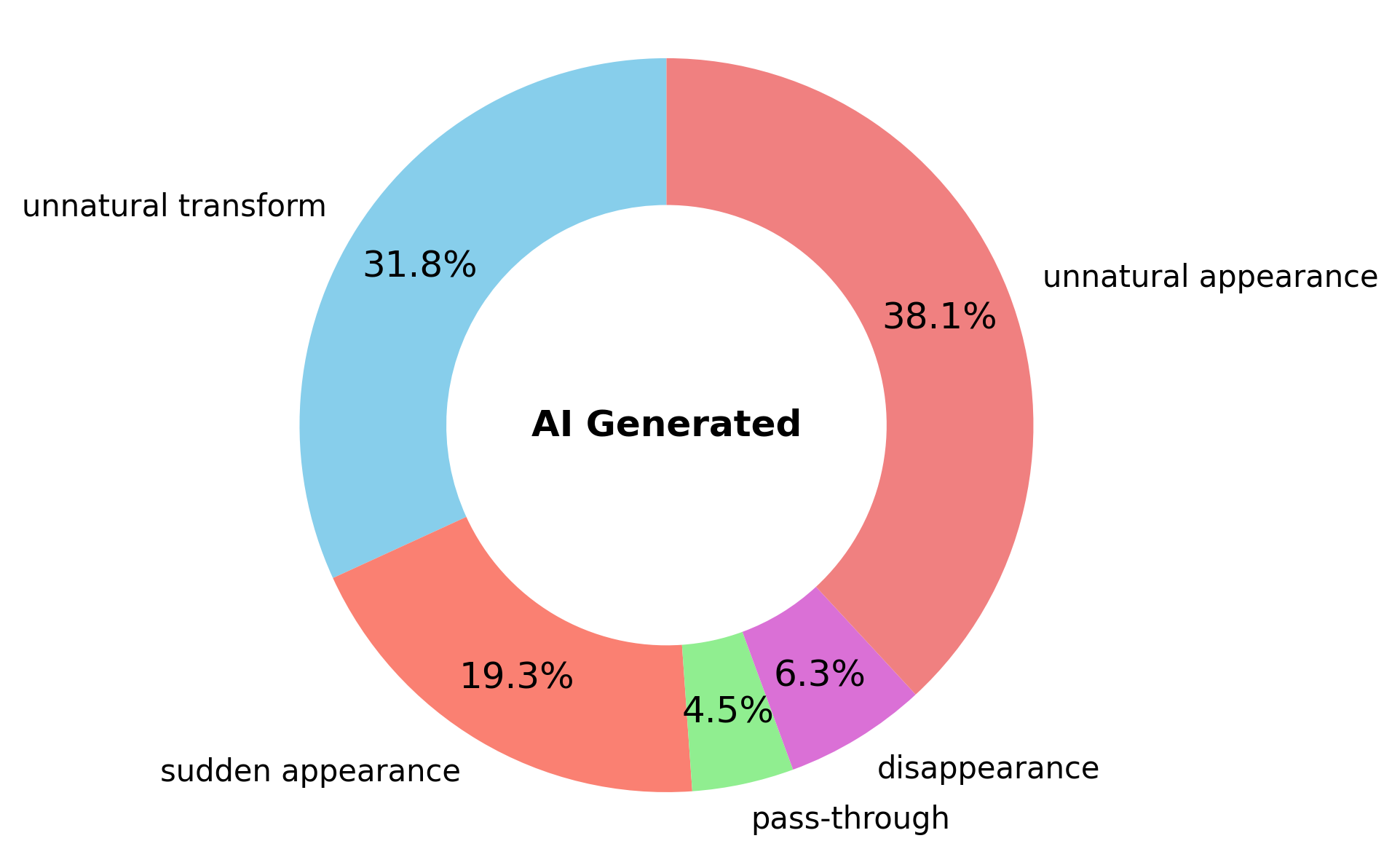}
        \label{fig:figure1}
    \end{minipage}
    \hfill
    \begin{minipage}[b]{0.26\textwidth}
        \centering
        \includegraphics[width=\textwidth]{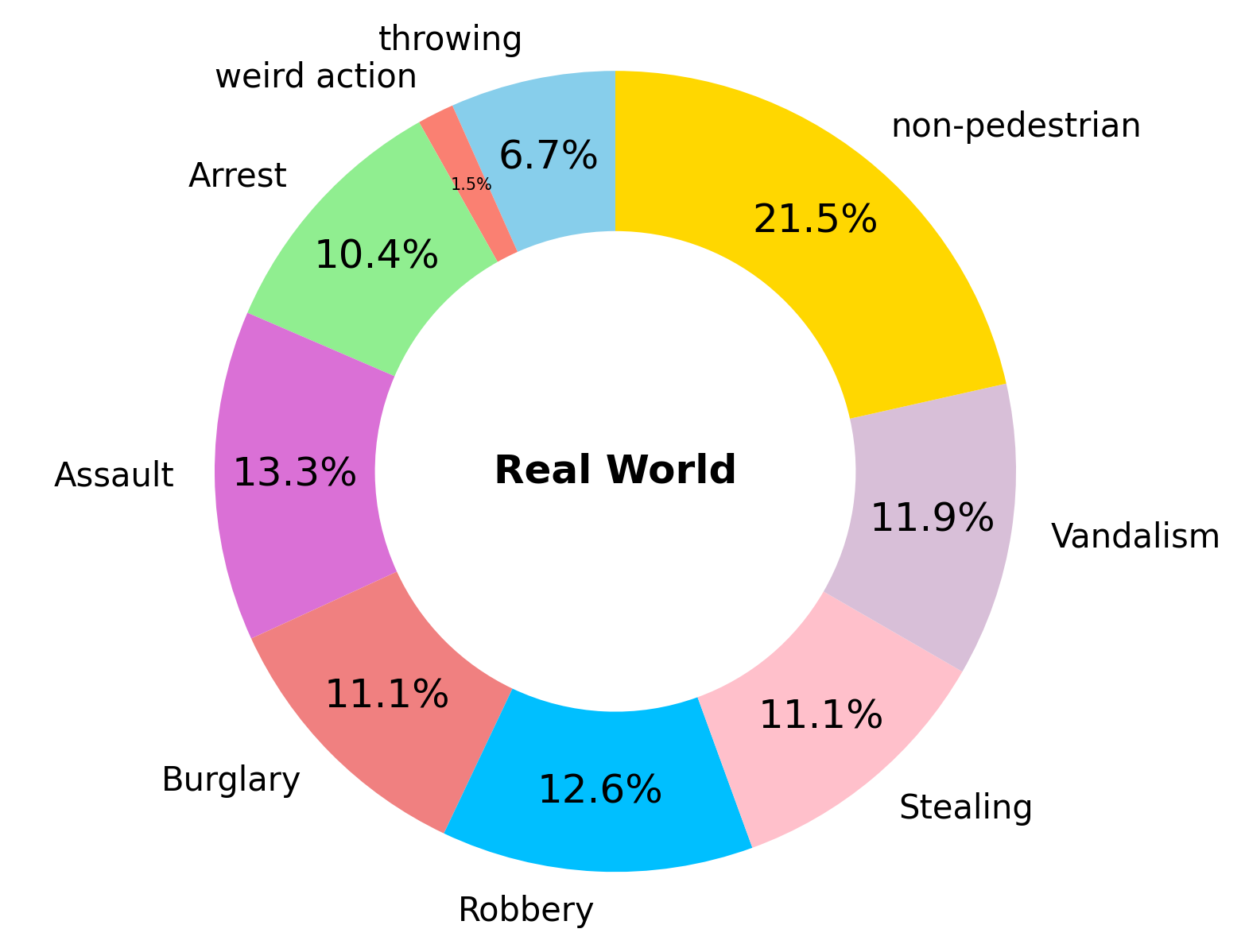}
        \label{fig:figure2}
    \end{minipage}
    \hfill
    \begin{minipage}[b]{0.4\textwidth}
        \centering
        \includegraphics[width=\textwidth]{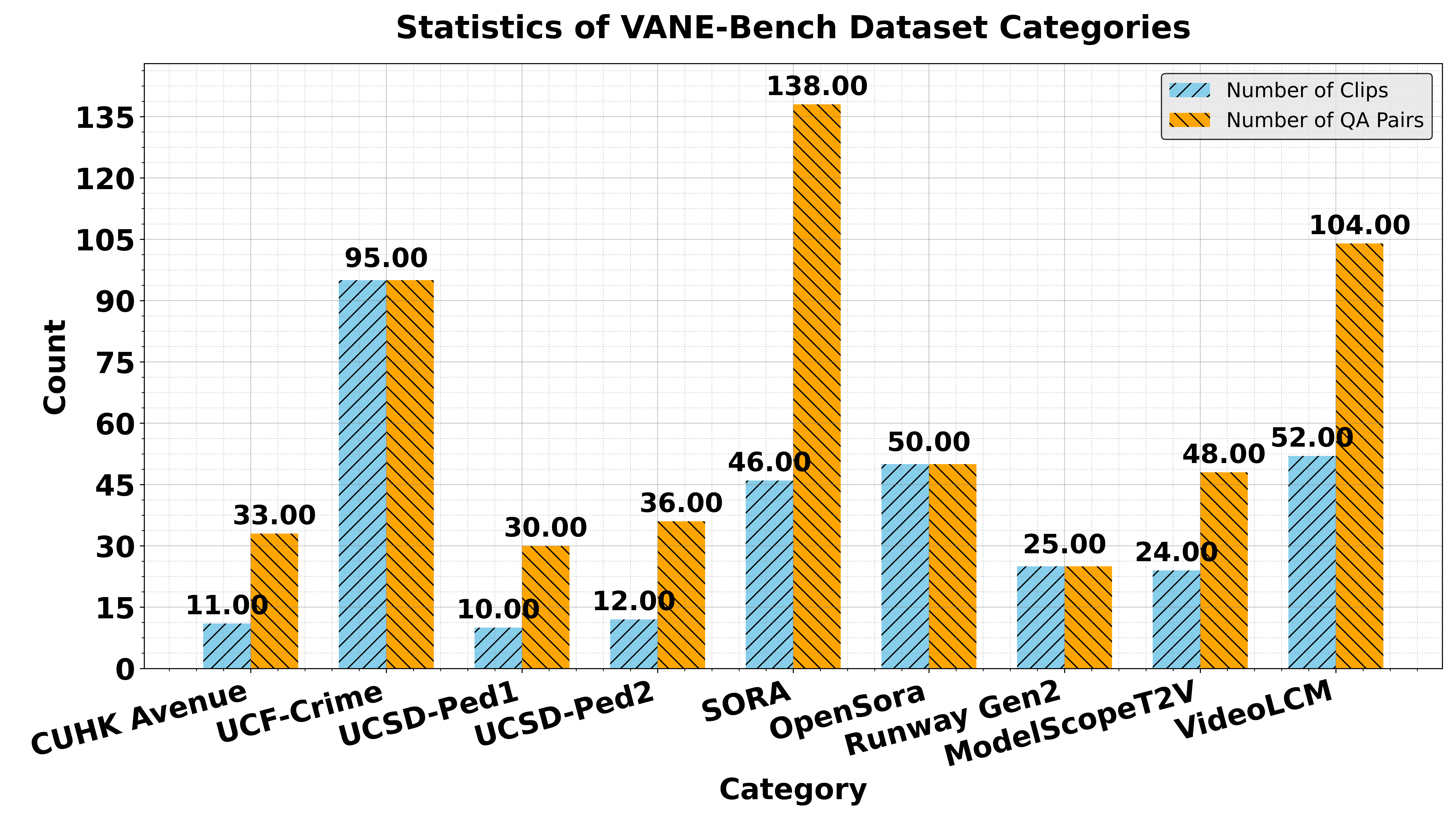}
        \label{fig:figure3}
    \end{minipage}
    \caption{VANE-Bench dataset statistics: \textbf{Left and Middle: }Composition and type of anomalies present in AI-generated and real-world videos. \textbf{Right: }Number of samples and QA pairs present in each type of video dataset. }
    \label{fig:dataset-stats}
    \vspace{-0.3cm}
\end{figure*}

The VANE-Bench dataset encompasses a variety of categories derived from both real-world surveillance footage and AI-generated video clips. Each category represents a distinct source and type of video anomaly. Below, we detail the different categories included in the dataset:

\noindent
\textbf{Real-World anomalies:} The videos with these anomalies are sourced from several established real-world anomaly datasets, encompassing diverse anomaly types. The distribution of these anomalies is depicted in Fig. \ref{fig:dataset-stats} (middle). Fig. \ref{fig:dataset-stats} (right) provides the total number of anomaly clips along with corresponding QA pairs for each dataset in this category. Detailed descriptions of each dataset within this category follow below.
    \begin{enumerate}
        \item \textbf{CUHK Avenue~\cite{cuhk-avenue}:} This category consists of 11 video clips with 33 associated question-answer (QA) pairs. The clips capture anomalous events in a campus environment, which shows individuals commuting in a university campus, and walking in and out of buildings. \textbf{Anomaly types. }The anomalies include unusual pedestrian behavior like randomly throwing bags and papers or performing weird actions or dance moves.
        \item \textbf{UCF-Crime~\cite{ucf-crime}:} Comprising 95 video clips with 95 QA pairs, this category includes real-world surveillance footage. \textbf{Anomaly types.} The videos depict various criminal activities, such as arrest, assault, burglary, robbery, stealing, and vandalism.
        \item \textbf{UCSD-Ped1~\cite{ucsd-ped}:} This category contains 10 video clips with 30 QA pairs. The videos focus on pedestrian walkways.  The Ped1 dataset is captured by a camera facing perpendicular to the road. \textbf{Anomaly types. } The anomalous events are due to the presence of non pedestrian entities (i.e. bikers, skaters, small carts, and wheelchairs) in the walkways.
        \item \textbf{UCSD-Ped2~\cite{ucsd-ped}:} Similar to UCSD-Ped1, this category includes 12 video clips with 36 QA pairs. In contrast with Ped1, the Ped2 dataset uses camera which is parallel to the road. \textbf{Anomaly types. }Abnormal events are due to non pedestrian entities in the walkways including bikers, skaters, small carts, and people walking across a walkway.
    \end{enumerate}
\textbf{AI-Generated anomalies:} The videos with these anomalies are obtained from various closed-source, and open-source text-to-video diffusion models. The anomalies in these clips are usually subtle, and hard to detect, which makes our VANE-Bench benchmark challenging. \textbf{General anomaly types: }The anomalies include the sudden appearance of objects, the unnatural transformation of solid physical objects, the disappearance of objects, objects passing through other solids, and unnatural appearance of objects (i.e., distorted and deformed facial features, or other unnatural appearance like presence of extra fingers). The distribution of these anomalies in the dataset is shown in Fig. \ref{fig:dataset-stats} (left), and statistics about the number of clips and corresponding QA pairs are presented in Fig. \ref{fig:dataset-stats} (right). Below, we describe the type of video samples in this category.
    \begin{enumerate}
        \item \textbf{SORA~\cite{sora}:} This category consists of 46 video clips with 138 QA pairs. The video clips are generated using SORA, a state-of-the-art AI text-to-video model. Due to the high quality and almost realistic-looking videos generated by SORA, it becomes quite difficult to accurately identify the inconsistencies or anomalies present in the videos.
        \item \textbf{OpenSora~\cite{open-sora}:} With 50 video clips and 50 QA pairs, this category features AI-generated videos from the open-source version of SORA.
        \item \textbf{Runway Gen2~\cite{gen2-runway}:} This category includes 25 video clips with 25 QA pairs created using a commercial text-to-video AI model.
        \item \textbf{ModelScopeT2V~\cite{wang2023modelscope}:} This category comprises of 24 video clips with 48 QA pairs, leveraging the video diffusion model trained by \cite{wang2023modelscope} to produce videos from text captions. The videos were generated with 50 diffusion steps with 16 fps.
        \item \textbf{VideoLCM~\cite{wang2023videolcm}:} This category features 52 video clips with 104 QA pairs, generated using latent consistency models \cite{wang2023videolcm} designed to create videos with high variability and with less latency. We used 20 diffuson steps to generate the videos with 16 fps. The videos were further post-processed by an LCM model trained on higher resolution videos to obtain better quality videos.
    \end{enumerate}

By including a wide range of video sources and anomaly types, the VANE-Bench dataset provides a comprehensive benchmark for evaluating the capabilities of large multi-modal models in video anomaly detection.

\subsection{Constructing VANE-Bench}
\label{sec:data-collect}

\begin{figure*}[!t]
    \centering
    \includegraphics[width=0.9\textwidth]{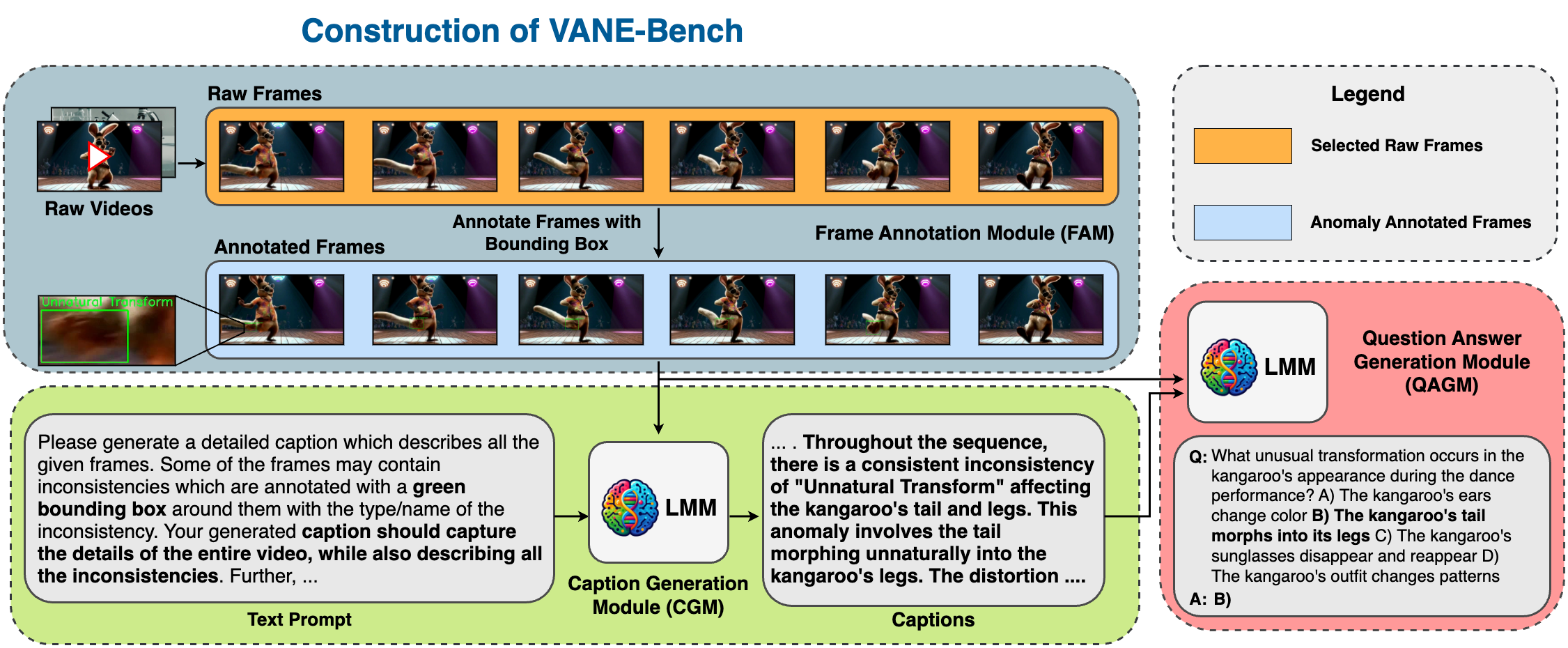}
    \caption{Flow diagram showing the semi-automatic construction process of our VANE-Bench dataset. The entire process can be divided into 3 interconnected stages/modules, i.e., i. Frame Annotation Module (FAM), ii. Caption Generation Module (CGM), iii. Question Answer Generation Module (QAGM).}
    \label{fig:VANE-Bench-construction}
\end{figure*}

Fig.~\ref{fig:VANE-Bench-construction} describes the construction process of the VANE-Bench dataset. Since the synthetic AI-generated videos from state-of-the-art models like SORA~\cite{sora} have subtle and hard-to-detect inconsistencies, we require high-quality captions describing all of the specific inconsistencies present in the given video. Our pipeline first annotates the anomalies using the frame annotation module (FAM). The caption-generating module (CGM) then utilizes these annotations to produce captions, followed by the question-answer generation module (QAGM), creating QA pairs based on the annotated frames and captions. Annotating the clips before caption generation is crucial for focusing the model on the specific anomaly regions in the video \cite{shtedritski2023does,zhang2023gpt4roi,yang2023set}. Without annotations, the CGM often fails to reference the anomalies in the captions, as demonstrated in Sec. \ref{appendix:fam-importance} of supplementary material. We briefly describe all the three stages involved in the semi-automatic dataset construction pipeline below.

\subsubsection{Frame Annotation Module (FAM)}
As described in Sec.~\ref{sec:data-category}, we first collect raw videos from existing VAD datasets like CUHK Avenue~\cite{cuhk-avenue}, UCF-Crime~\cite{ucf-crime}, UCSD-Ped~\cite{ucsd-ped}, and also add additional challenging AI generated videos to the mix. For the VAD datasets, the bounding box annotations were already provided for a subset of the videos from these datasets. Thus, we only annotate the anomalies present in the AI-generated videos. In this stage, we first break down the raw videos into their constituent image frames. Second, we select and filter 10 consecutive frames from the video that contain the inconsistency. We annotate these selected frames with a bounding box mentioning the type of inconsistency. We consider the following inconsistency types: `\texttt{Sudden Appearance}', `\texttt{Unnatural Transform}', `\texttt{Disappearance}', `\texttt{Pass-through}, and `\texttt{Unnatural Appearance}'. Fig.~\ref{fig:VANE-Bench-construction} shows the annotated `\texttt{Unnatural Transform}' inconsistency affecting the kangaroo's legs and tails.

\subsubsection{Caption Generation Module (CGM)}
The second stage of our data collection process involves the Caption Generation Module (CGM), which uses the annotated video frames from FAM to generate a high-quality and detailed caption which describes the inconsistency, along with the general events in the video. To generate the caption, we design a specialised custom prompt (Sec.~\ref{sec:cgm-prompt}), and use the recently released GPT-4o~\cite{gpt4-o} LMM, which has shown both impressive performance gains and cost savings. Thus, GPT-4o model takes in our custom prompt, along with the annotated frames to generate the descriptive video caption as shown in Fig.~\ref{fig:VANE-Bench-construction}.

\subsubsection{Question Answer Generation Module (QAGM)}
The final stage of our VANE-Bench construction process involves using the generated caption from CGM, and the annotated frames from FAM to output the final high-quality, and challenging Question and Answer (QA) pairs. We create another custom prompt (Sec.~\ref{sec:qagm-prompt}) which we pass to the GPT-4o model, along with caption, and the annotated frames as input to generate the QA pairs. The selected raw frames containing the inconsistency, and their corresponding generated QA pairs form our VANE-Bench dataset.

\section{Experiments and Results}
\label{sec:exp-and-results}

\begin{table*}[!t]
    \small \centering
 \setlength{\tabcolsep}{3pt}
    \scalebox{0.85}{
\begin{tabular}{l | ccccccc|cc }
\toprule
Benchmark Category  & 
 \rotatebox{45}{Video-LLaMA} &
 \rotatebox{45}{VideoChat}  & 
 \rotatebox{45}{Video-ChatGPT}  & 
 \rotatebox{45}{Video-LLaVA}  &
 \rotatebox{45}{MovieChat} &
 \rotatebox{45}{LLaMA-VID} &
 \rotatebox{45}{TimeChat} &
 \rotatebox{45}{Gemini-1.5 Pro} &
 \rotatebox{45}{GPT4o} \\
\midrule

   SORA & 11.59 & 10.74 & 26.47 & 10.86 & 8.69 & 7.97 & 21.73 & 51.45 & 55.80 \\
\midrule
  OpenSORA & 18.00 & 28.00 & 22.00 & 18.00 & 10.00 & 14.00 & 26.00 & 84.00 & 68.00 \\
\midrule
   Runway Gen2 & 16.00 & 4.00 & 12.00 & 16.00 & 1600 & 20.00 & 28.00 & 28.00 & 40.00 \\
\midrule
  VideoLCM & 10.57 & 17.64 & 18.26 & 19.23 & 14.42 & 19.23 & 22.11 & 49.04 & 50.96 \\
\midrule
 Modelscope-T2V & 10.41 & 20.83 & 16.66 & 16.66 & 6.25 & 14.58 & 20.83 & 75.00 & 64.58 \\
\midrule
 Avenue & 30.00 & 32.25 & 39.39 & 3.03 & 18.18 & 27.27 & 24.20 & 100.00 & 84.85 \\
\midrule
 UCFCrime & 9.47 & 11.57 & 31.57 & 10.52 & 18.51 & 15.78 & 7.30 & 76.84 & 83.16 \\
\midrule
 UCSD-Ped1 & 16.66 & 13.33 & 40.00 & 2.77 & 6.66 & 6.66 & 27.58 & 96.67 & 93.33 \\
\midrule
 UCSD-Ped2 & 5.55 & 13.88 & 19.44 & 6.06 & 11.11 & 19.44 & 11.11 & 94.44 & 86.11 \\
\bottomrule
\end{tabular}
}
    \caption{ Evaluation results of Video-LMMs across different types of video samples on the VANE benchmark. We present results for both open-source and closed-source models. The first five rows show results on AI-generated videos and last four contain results on real world anomaly datasets.}
    \label{tab:1}
\end{table*}

\textbf{Video-LMMs.} We evaluate the anomaly detection and comprehension capabilities of both open-source and closed-source models. Among the open-source models, we evaluate 7 recent Video-LMMs, including Video-LLaVA \cite{lin2023video}, TimeChat \cite{ren2023timechat}, MovieChat \cite{song2023moviechat}, LLaMA-ViD \cite{li2023llama}, VideoChat \cite{li2023videochat}, Video-ChatGPT \cite{maaz2023video}, and Video-LLaMA-2 \cite{zhang2023video}. For evaluating closed-source models, we use Gemini-1.5 Pro \cite{Gemini} and GPT-4o \cite{gpt4-o}.

\vspace{-0.1cm}

\noindent
\textbf{Evaluation Protocol.} For the evaluation of Gemini and GPT-4o, we utilize their respective official APIs, with each model receiving 10 video frames as input. The 10 frames are selected in a manner that encompasses all or the majority of the inconsistencies present in the video. In cases where an anomaly spans a longer duration, we sample multiple sets of 10 frames to ensure comprehensive coverage. As GPT-4o does not inherently support videos, we input the video clips as 10 frames to the GPT API, accompanied by the corresponding Visual Question-Answering (VQA) query. For each model under assessment, we generate responses to the questions independently and without retaining the conversation history. Few models, such as Moviechat, output hallucinated responses when instructed to answer the query. In such cases, we consider the hallucinated responses as incorrect answers due to the inability of the model to comprehend the situation in the video. 
\vspace{-0.1cm}

\noindent
\textbf{Evaluation metric.} For the evaluation results of the Video-LMMs on our proposed VANE-Bench benchmark, we employ the standard VQA accuracy measure, which assigns a score of 1 to each correct answer and a score of 0 to each incorrect answer.

\subsection{Main Evaluation Results} 

\subsubsection{Evaluation on Video-LLMs}
\textbf{AI-Generated anomalies. }The AI-generated videos in our dataset are derived from five distinct models: SORA, OpenSORA, Runaway Gen-2, VideoLCM, and Modelscope-T2V. In the majority of these videos, the anomalies are subtle and not readily apparent, even to the human eye. As previously stated in section \ref{sec:data-category}, the synthetic anomalies can manifest in five different forms. As shown in Table \ref{tab:1}, the performance of open-source models in detecting anomalies in these videos is subpar. Although closed-source models outperform their open-source counterparts, their overall comprehension and detection of anomalies in the videos remain inadequate. This indicates that even robust closed-source models encounter difficulties in identifying subtle anomalies within the videos.

\noindent\textbf{Real-world anomalies. }Our real-world anomaly datasets benchmark, as discussed in section \ref{sec:data-category}, comprises four real-world datasets and focuses on detecting crime-related irregularities, atypical pedestrian behavior, and unusual events. These anomalies are prevalent in real-world scenarios. In our analysis, we find that open-source models encounter difficulties in locating and identifying these anomalies. As shown in Table \ref{tab:1}, these models perform poorly on these datasets. Conversely, we observe that closed-source models excel at detecting such real-world anomalies, indicating that they can effectively differentiate between unusual events in real-world scenarios. This can be attributed to the fact that these models are trained on a vast amount of existing real-world, internet-scale data.

We provide results on additional latest Video-LMMs in Section \ref{appendix:quavtitative-sub} of Supplementary.

\begin{figure}
  \begin{center}
    \includegraphics[width=0.9\columnwidth]{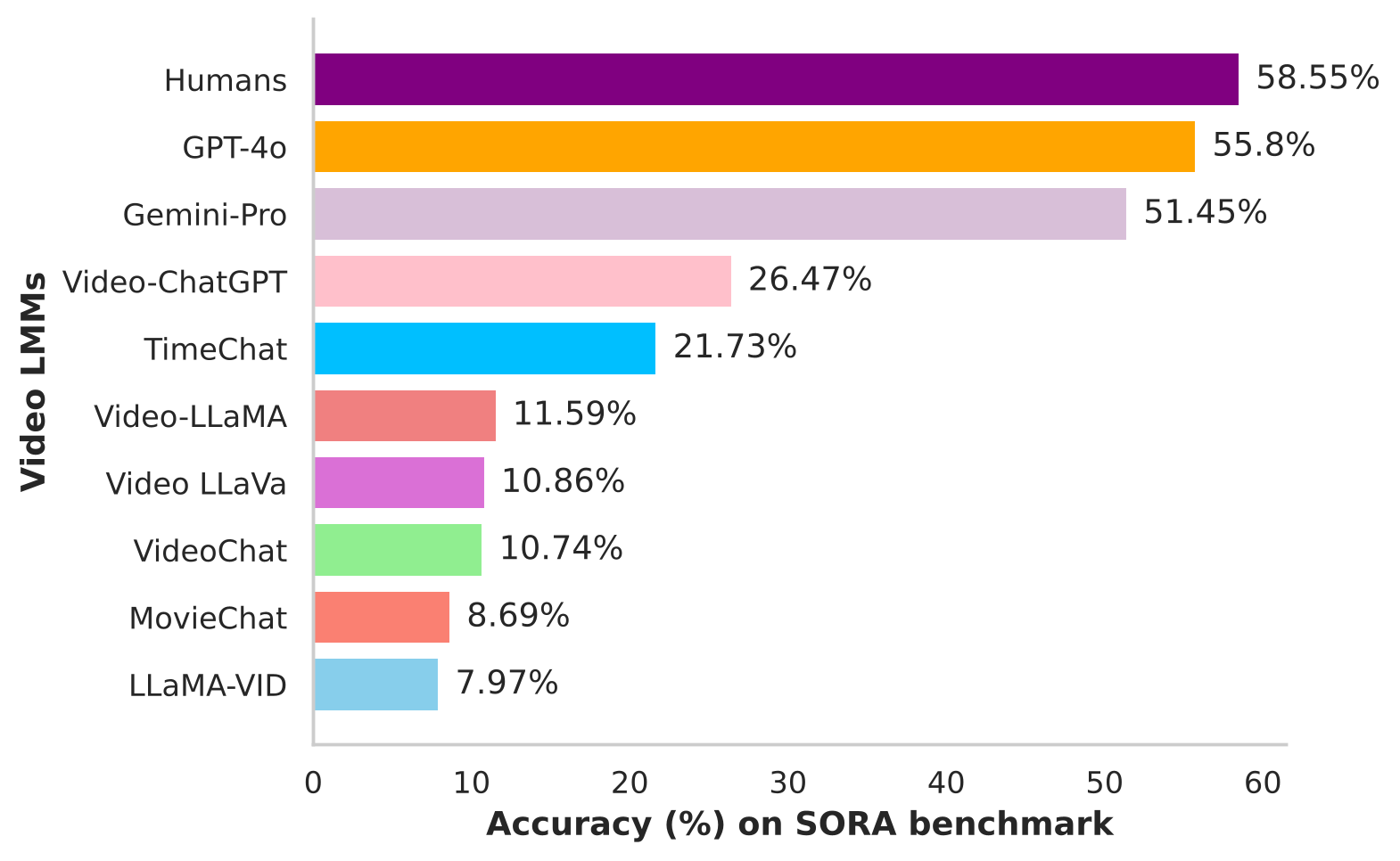}
  \end{center}
  \vspace{-1em}
\caption{\textbf{Human vs Video-LMMs' performance on SORA.} Performance comparison of humans vs Video-LMMs on VQA task of detecting anomalies in SORA dataset. We find that closed-source Video-LMMs perform comparably to humans while open-source Video-LMMs struggle to detect subtle anomalies.}
\vspace{-1.5em}
 \label{fig:bar-plot-sora-human}
\end{figure}

\begin{figure*}[!t]
    \centering
    \includegraphics[width=0.95\textwidth]{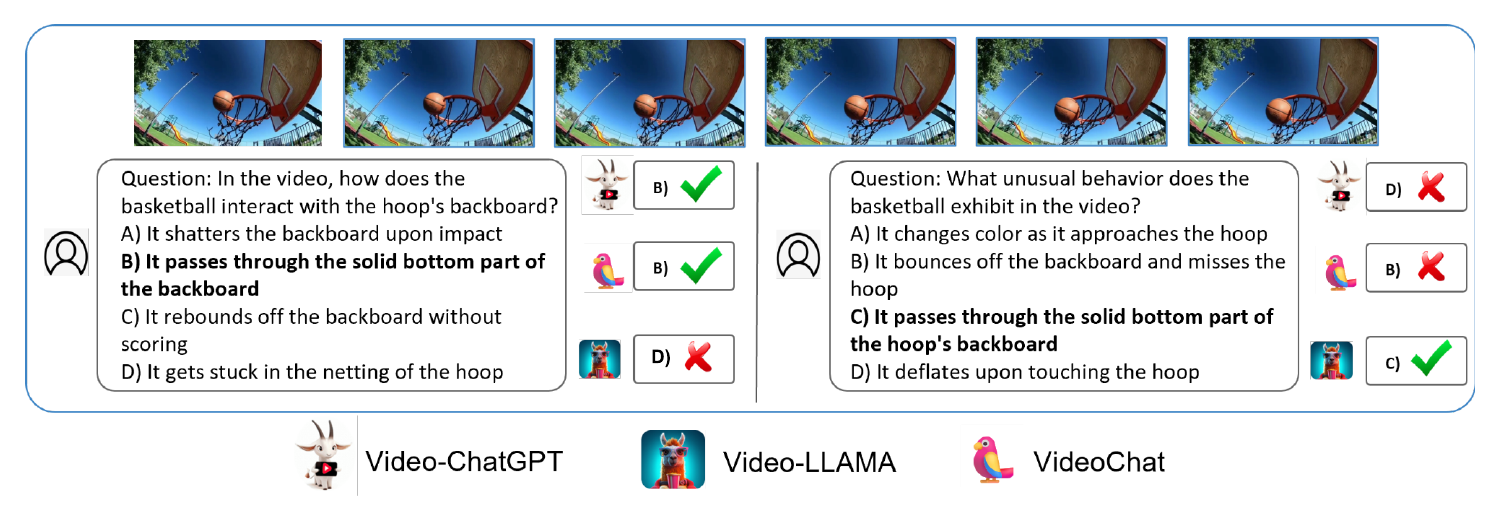}
    \caption{Inconsistency in Predictions: \textbf{Left:} Video-ChatGPT and VideoChat predict accurately, while Video-LLAMA selects incorrectly. \textbf{Right:} With a rephrased query, 
 predictions shift. Video-ChatGPT and VideoChat err, whereas Video-LLAMA predicts correctly. This indicates the sensitivity of Video-LMMs towards query rephrasing.}
 \label{fig:inconsistent-pred}
 \vspace{-0.2cm}
\end{figure*}

\subsubsection{Human Evaluation}
\label{subsec:human-eval}
We conducted a human evaluation on SORA-generated videos, which contain subtle and challenging anomalies that are difficult for humans to detect (see Fig. \ref{fig:intro-fig} top row) in a single viewing.  Moreover, most of the video clips contain a multitude of foreground and background characters and elements, which makes it difficult for humans to focus on the inconsistencies within the short time frame. Some of the questions also specifically inquire about inconsistencies present in the background characters of the clips rather than the foreground ones. To ensure fairness, our human evaluation was conducted under a set of rules, which include showing all 10 frames of the video to the human evaluator only once, followed by the question. Our human evaluation comparisons are presented in Fig \ref{fig:bar-plot-sora-human}. While humans outperform open-source models in detecting these subtle anomalies, their performance remains sub-optimal. This indicates that, with the advancements in video generation techniques, there is a pressing need for more sophisticated and effective Video-LMMs capable of assisting in the detection of such challenging cases capable of evading human eyes as well.

\subsection{Additional Analysis}
\label{additional-analysis}
\noindent \textbf{Inconsistencies in Predictions. }We find that, in the majority of cases, open-source Video-LMMs generate different results when prompted to answer the same query multiple times. Fig. \ref{fig:inconsistent-pred} illustrates a sample example where the same questions were posed twice to the corresponding Video-LMMs, yielding different responses. In some instances, the answers generated by the Video-LMMs in both rounds were dissimilar and incorrect. However, we also found cases where Video-LMMs initially produced the correct answer, followed by an incorrect answer to the same query, albeit phrased slightly differently. This suggests that the majority of these open-source Video-LMMs struggle to comprehend the same query when presented in a different manner, leading to inconsistent and paradoxical predictions. In contrast, closed-source Video-LMMs are less prone to such inconsistent predictions and consistently produce the same output for the same queries, regardless of how they are phrased, indicating a superior comprehension of language. Refer to supplementary Section \ref{appendix:pred-inconsistency} for additional results.

\noindent \textbf{Performance Analysis on SORA anomalies.} The overall performance of open-source Video-LMMs on anomaly categories in synthetically generated SORA videos is subpar. To gain further insights, as depicted in Figure \ref{fig:intro-accuracy-all-datasets} (left), all open-source Video-LMMs exhibit less than 10\% accuracy in detecting the ``disappearance" anomaly, indicating that this particular type is the most difficult to identify for the majority of Video-LMMs. Among the open-source models, Videochat demonstrates above par performance compared to its open-source counterparts on most anomaly types, with the exception of the "unnatural appearance" category, where Timechat outperforms it. The remaining models display a fluctuating trend, with accuracy levels ranging from extremely low to moderately low across all anomaly types. The closed source-models, on the other hand, demonstrate superior performance compared to open-source models across all anomaly types.

We provide more insights and discussions in Section \ref{Additional-Discussions-on-Experimental-Results} of Supplementary material.

\section{Conclusion}

We introduced VANE-Bench, a comprehensive benchmark for evaluating Video LMMs in VAD tasks, featuring real-world and AI-generated video clips. The AI-generated content, especially from advanced models like SORA, includes subtle inconsistencies, making VANE-Bench particularly challenging. Our evaluation of nine recent Video-LMMs on VANE-Bench shows significant gaps in detecting video anomalies, with even robust closed-source models struggling with nuanced discrepancies. Human assessments on SORA-generated videos confirm these subtle anomalies are challenging to identify, highlighting the need for advanced Video-LMMs. VANE-Bench is vital for advancing Video-LMMs in anomaly detection. As high-fidelity AI-generated content rises, our benchmark is crucial for developing models to identify subtle inconsistencies, aiding in the fight against misinformation and deepfakes. We hope VANE-Bench will guide future research to enhance the robustness and capability of Video-LMMs in this critical area.

\section{Limitations}
\label{limitations}
Our VANE-Bench is the first benchmark for evaluating Video-LMMs on anomalous videos from both AI-generated and real-world sources. While we have done our best to ensure a high-quality evaluation of these Video-LMMs, certain limitations still manifest.

Our Question-answer pairs are designed to have 4 options. We design the instruct prompt to ensure that each Video-LMM outputs one out of 4 options. However, in some instances, the model outputs a hallucinated response and does not follow the instructions. As a result, we employ a post-response human-based filtration process, which involves an exhaustive verification
and rectification of these errors. In our current setup, we mark these cases as wrong. We believe that future Video-LLMs will be more aligned with human intent and will follow human instructions appropriately.

Additionally, the video samples from the SORA are limited in VANE-Bench. This is due to the fact that SORA model is not open-source yet, hence we rely on publicly available samples of SORA for evaluation.

\bibliography{main}

\clearpage
\appendix

\noindent\begin{LARGE} \textbf{Appendix} \vspace{4mm} \end{LARGE}

In the following sections, we provide additional information for the paper: \textbf{VANE-Bench: Video Anomaly Evaluation Benchmark
for Conversational LMMs}. The contents are organized in the following order. 

\begin{itemize}
% [topsep=0pt,itemsep=-1ex,partopsep=1ex,parsep=1ex]
\item Additional Findings and Results (Appendix~\ref{appendix:qualitative_results})
\item Additional Results on Prediction Inconsistency (Appendix~\ref{appendix:pred-inconsistency})
\item Importance of Frame Annotation Module (FAM) (Appendix~\ref{appendix:fam-importance})
\item Implementation Details (Appendix~\ref{appendix:implementation-details})
\item Distribution of VANE-Bench dataset (Appendix~\ref{appendix:vane-bench-distribution})
    
\end{itemize}

\section{Additional Findings and Qualitative Results}
\label{appendix:qualitative_results}

\subsection{Additional Quantitative Results}
\label{appendix:quavtitative-sub}

Video-LMMs are fewer in number compared to image-based multi-modal models, which limits the range of options available for evaluation. Given this scarcity, we selected 7 open-source and 2 closed-source LMMs that are currently among the most widely used. To ensure that our benchmark remains representative, we have also included additional results from other latest open-source LMMs: \cite{ataallah2024minigpt4,ataallah2024goldfishvisionlanguageunderstandingarbitrarily,li2024llava}, as shown in Table \ref{tab:A1}. Our findings reveal that open-source models still lag behind their closed-source counterparts in performance, indicating that simply adding more models wouldn't necessarily improve the overall representativeness of our benchmark. Our selected set, which includes both open-source and closed-source models, is already comprehensive, featuring state-of-the-art models like GPT-4o and Gemini-1.5 Pro. Given the limited number of Video-LMMs available, our reliance on this specific set of models is justified, as it accurately represents the current landscape of Video-LMM capabilities.

\begin{table}[!t]
    \small \centering
 \setlength{\tabcolsep}{3pt}
    \scalebox{0.85}{
\begin{tabular}{l | ccc}
\toprule
Benchmark Category  & 
LLaVA-NeXT &
 MiniGPT4-Video  & 
 Goldfish \\
\midrule

   SORA & 11.59 & 10.74 & 26.47 \\
\midrule
  OpenSORA & 18.00 & 28.00 & 22.00 \\
\midrule
   Runway Gen2 & 16.00 & 4.00 & 12.00 \\
\midrule
  VideoLCM & 10.57 & 17.64 & 18.26 \\
\midrule
 Modelscope-T2V & 10.41 & 20.83 & 16.66  \\
\midrule
 UCFCrime & 9.47 & 11.57 & 31.57 \\
\midrule
 UCSD-Ped1 & 16.66 & 13.33 & 40.00  \\
\midrule
 UCSD-Ped2 & 5.55 & 13.88 & 19.44 \\
\bottomrule
\end{tabular}
}
\caption{\small Evaluation results of additional latest Video-LMMs across different types of video samples on the VANE benchmark. We present results for both open-source and closed-source models. The first five rows show results on AI-generated videos and last three contain results on real world anomaly datasets.}
    \label{tab:A1}
\end{table}

\subsection{Qualitative Results}
\label{appendix:qualitative-sub}

\begin{figure*}[!t]
    \centering
    \includegraphics[width=1.0\textwidth]{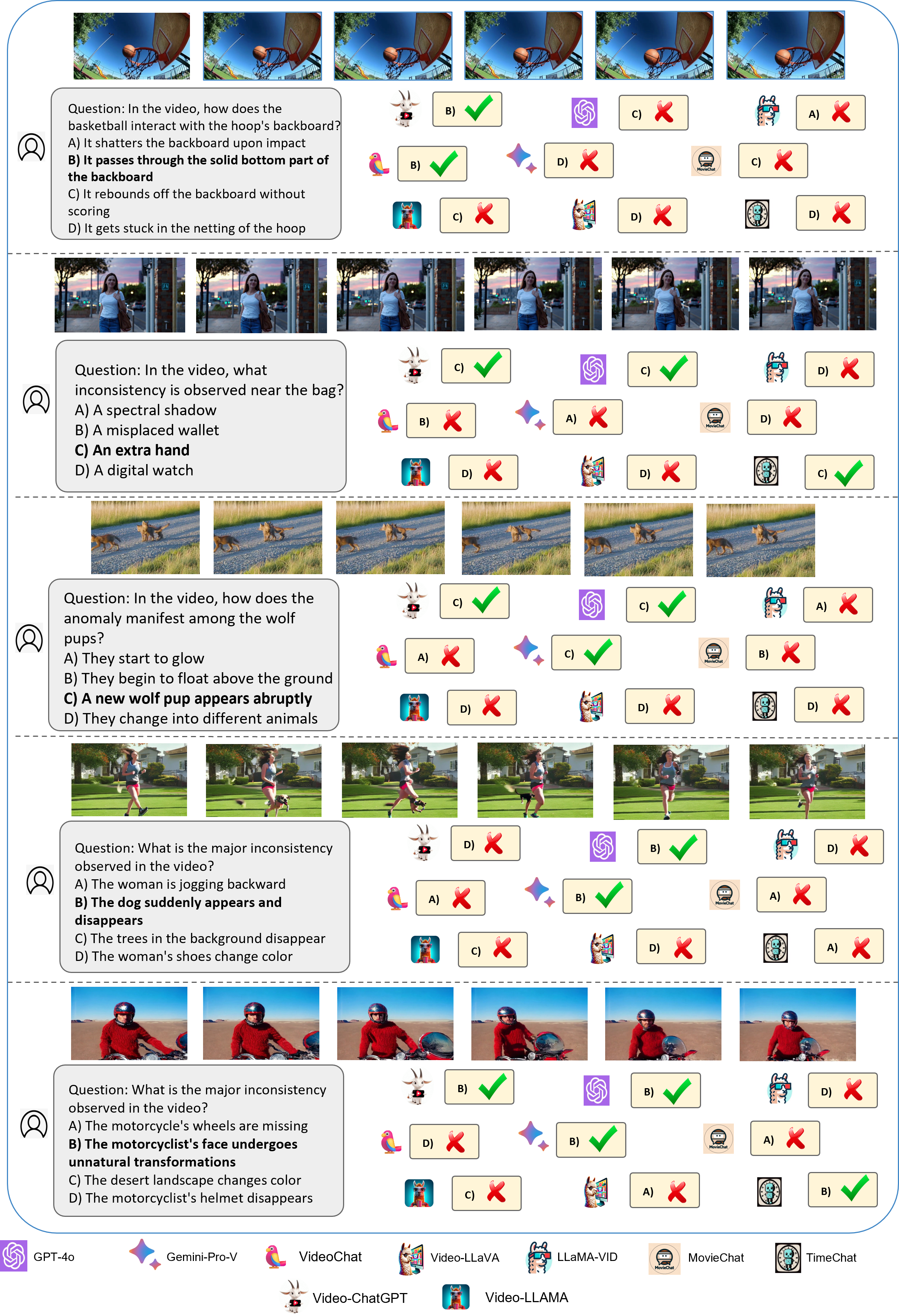}
    \caption{\textbf{Qualitative examples: }Figure shows the response of Video-LMMs to the VQA task of detecting anomalies in the video. The correct answer is written in bold in the user query. We find that majority of Video-LMMs struggle to answer the questions correctly.}
    \label{fig:vane-appendix-qualitative}
\end{figure*}

In Fig. \ref{fig:vane-appendix-qualitative}, we showcase the response of both open-source and closed-source Video-LMMs on anomalous video samples from our VANE-benchmark. The query to the Video-LMMs contains the video and a question with multiple options associated with the specific anomaly present in the video. The anomalies in Fig. \ref{fig:vane-appendix-qualitative} constitute pass through (first row), unnatural appearance (second row), sudden appearance (third row), disappearance (fourth row) and unnatural transformation (fifth row).

\subsection{VANE-Bench frequent instances}
\label{appendix:frequent-instances}

\begin{figure}[!h]
    \centering
    \includegraphics[width=1.0\columnwidth]{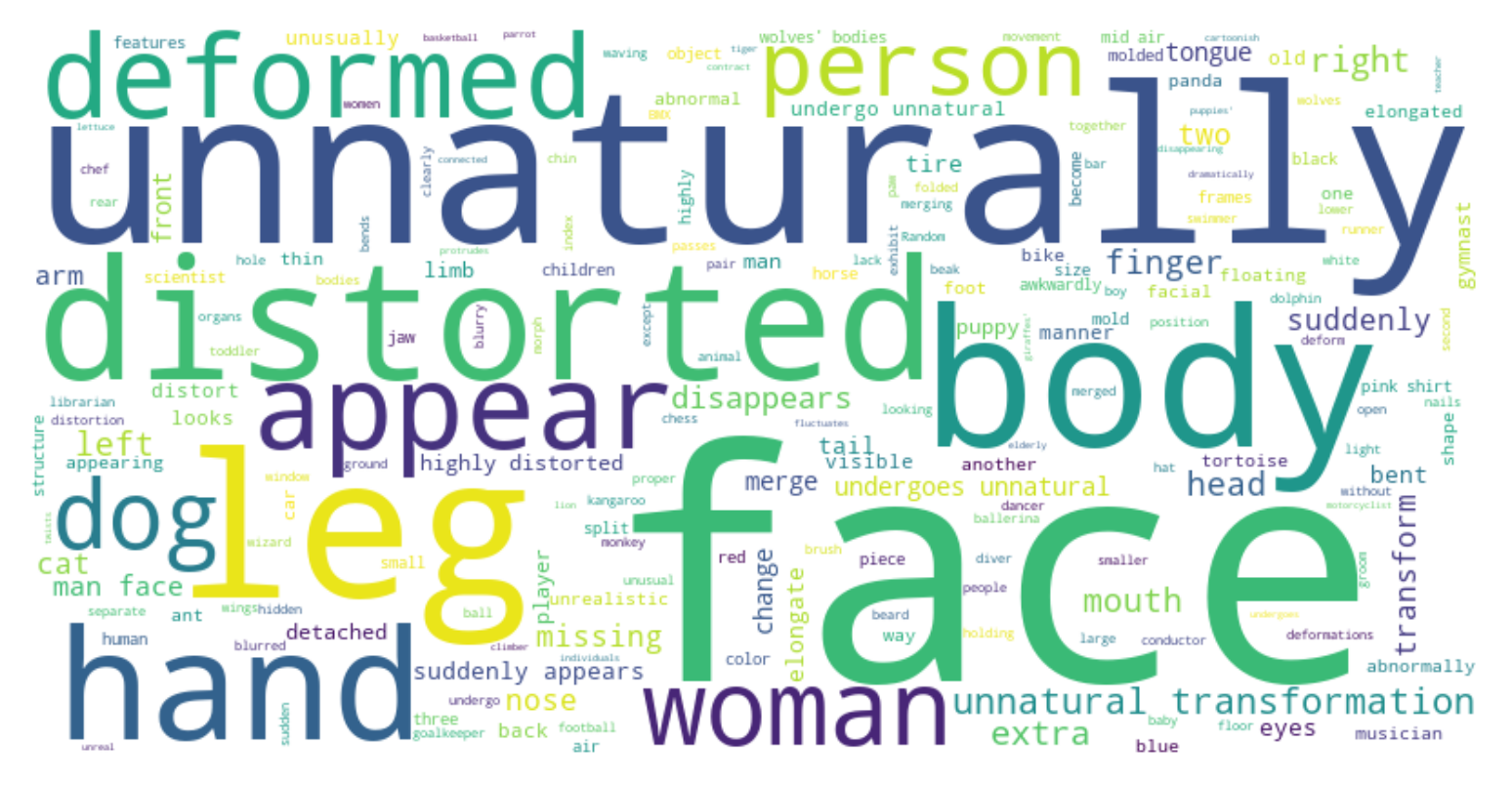}
    \caption{\textbf{Frequent keywords:  }
Illustration of the most frequent keywords in the correct option set of VANE benchmark. These keywords signify the objects or human attributes in the videos that are most likely to exhibit anomalous behavior }
    \label{fig:wordcloud}
\end{figure}

Figure \ref{fig:wordcloud} presents a word cloud visualization, highlighting the most frequently occurring keywords within the correct option set of the VANE-Benchmark dataset. These prominent words are indicative of objects or human attributes in the videos that are most likely to exhibit anomalous behavior. From the figure, the most frequently occurring keyword is "Face" which indicates that the synthetically generated videos most likely struggle to generate a perfect human face.

\subsection{Additional Discussions on Experimental Results}
\label{Additional-Discussions-on-Experimental-Results}
\noindent
\textbf{Per anomaly performance:} To give further insights, Figure \ref{fig:intro-accuracy-all-datasets} (left) of the main paper illustrates the performance of LMMs on each type of anomaly present in the AI-generated videos. We can observe that closed-source models like GPT-4o and Gemini-1.5 Pro consistently exhibit strong performance across all five anomaly categories compared to their open-source counterparts. This likely stems from their access to significantly larger training datasets and model parameters, allowing for a more robust understanding of visual anomalies. Conversely, open-source models exhibit fluctuating performance depending on the anomaly type. We also note that open-source models struggle, especially with the “disappearance” anomaly. We believe that it might be because of the fact that these models are trained on datasets focusing on the presence of objects and actions, and hence being more biased towards presence. Further, we believe that open-source models suffers from limited temporal reasoning capability and often use short-term mechanisms that limit their ability to track objects over time. The lack of datasets focusing on anomalies like “disappearance” also limits the model’s capability to detect such patterns.

\noindent
\textbf{Higher performance of some LMMs:} As seen in Table \ref{tab:1} and Figure \ref{fig:bar-plot-sora-human} of main paper, we notice that some open-source LMMs perform better than their counterparts. For instance, we notice Video-ChatGPT achieves higher performance compared to other open-source models. We believe that it might be because of the following two reasons: 1. Training Data: While most open-source models rely solely on web-scraped video captioning data, Video-ChatGPT incorporates a manually annotated video instruction dataset specifically designed for video understanding. This provides the model with a more direct and targeted learning experience, potentially enhancing its sensitivity to anomalies. 2. Two-Stage Training: Video-ChatGPT employs a two-stage training process involving both video-language pre-training and instruction tuning. This enables the model to first develop a strong understanding of general video semantics and then refine its ability to follow user instructions and reason about specific events within videos.

\section{Additional results on Prediction inconsistency}
\label{appendix:pred-inconsistency}
\begin{figure*}[!t]
    \centering
    \includegraphics[width=1.0\textwidth]{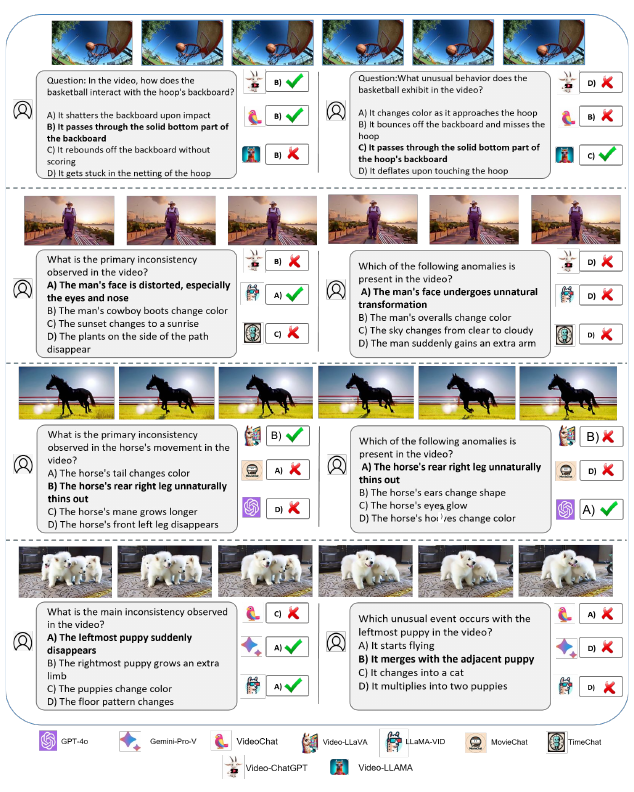}
    \caption{\textbf{Prediction Inconsistency: }Figure shows the response of Video-LMMs to the VQA task of detecting anomalies in the video. The correct answer is written in bold in the user query. We find that the majority of Video-LMMs struggle to answer the questions correctly.}
    \label{fig:appendix-pred-inconsistency}
\end{figure*}

As discussed in section \ref{additional-analysis} almost all Video-LMMs generate different results when prompted to answer the same query rephrased multiple times. While it is most common in open-source Video-LMMs, we found that closed-source Video-LMMs occasionally suffer from this problem as well. Fig. \ref{fig:appendix-pred-inconsistency} illustrates additional sample examples where the same questions (phrased slightly differently) were posed twice to the corresponding Video-LMMs, yielding different responses.

We find that, in the majority of cases, open-source Video-LMMs generate different results when prompted to answer the same query multiple times. Fig. \ref{fig:inconsistent-pred} illustrates a sample example where the same questions were posed twice to the corresponding Video-LMMs, yielding different responses. In some instances, the answers generated by the Video-LMMs in both rounds were dissimilar and incorrect. However, we also found cases where Video-LMMs initially produced the correct answer, followed by an incorrect answer to the same query, albeit phrased slightly differently. This suggests that the majority of these open-source Video-LMMs struggle to comprehend the same query when presented in a different manner, leading to inconsistent and paradoxical predictions. In contrast, closed-source Video-LMMs are less prone to such inconsistent predictions and consistently produce the same output for the same queries, regardless of how they are phrased, indicating a superior comprehension of language.

\section{Importance of Frame annotation module}
\label{appendix:fam-importance}
\begin{figure*}[ht]
    \centering
    \includegraphics[width=\textwidth]{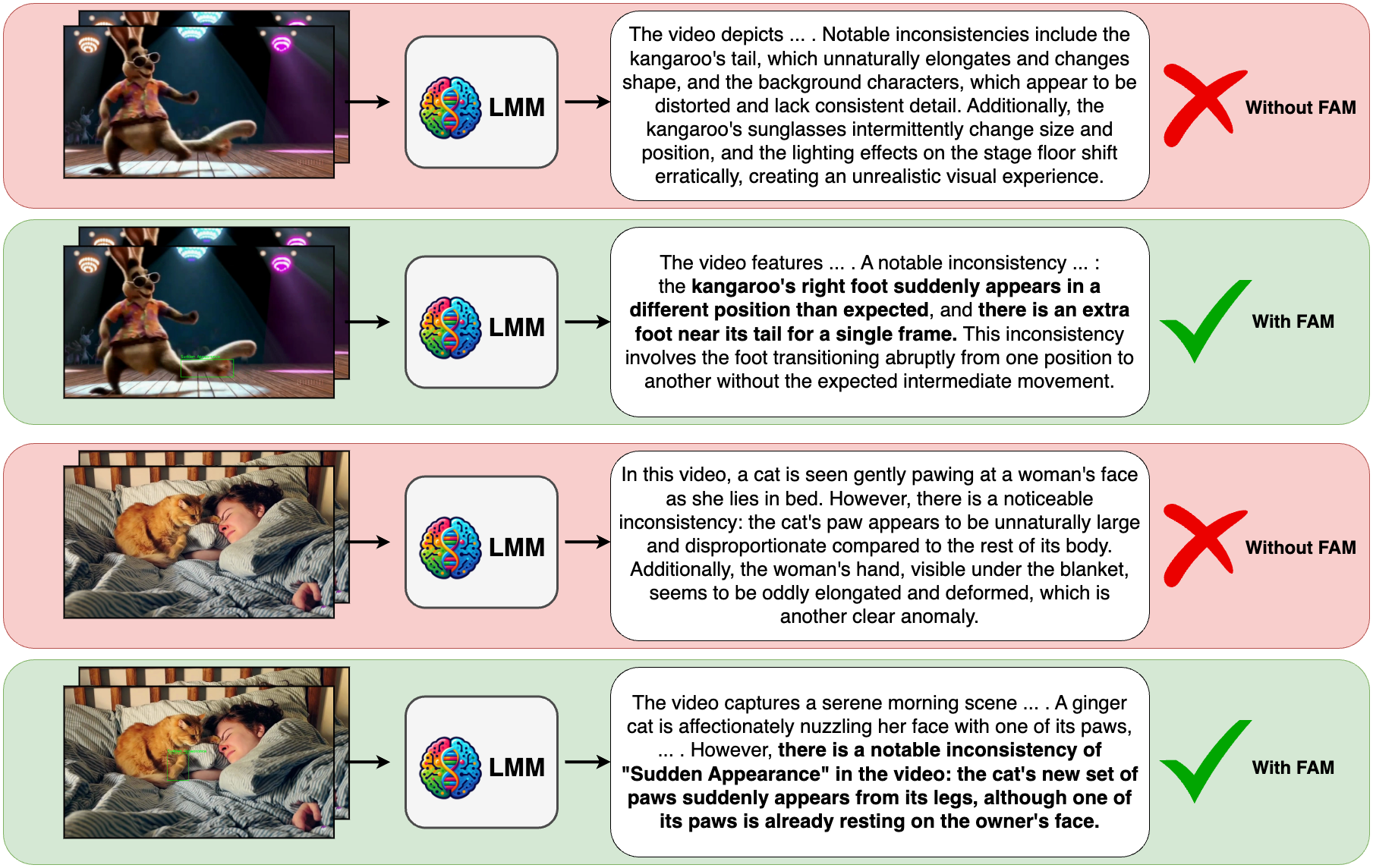}
    \caption{Example showcasing the importance of our Frame Annotation Module (FAM). We note that without FAM, the LMM responsible for generating the captions is not able to identify or describe the accurate anomaly present in the video. However, by providing the bounding box annotation for the inconsistency, we are able to ensure that the generated caption accurately describes the anomaly in the video.}
    \label{fig:fam-imp}
\end{figure*}

Since the video inconsistencies present in state of the art AI models like SORA are quite subtle, and hard to detect, our Frame Annotation Module (FAM) ensures that we are able to generate high-quality and accurate captions for these videos. As shown in Fig. \ref{fig:fam-imp}, without FAM, the generated caption is not able to describe the sudden appearance of the kangaroo's right foot near its tail. Further, the caption generated without our FAM is also not able to describe the extra set of paws that appear suddenly from the legs of the cat. Thus, FAM plays an important role in curating high-quality and accurate video captions.

\section{Implementation Details}
\label{appendix:implementation-details}
We use the official code of each open-source Video-LMM for evaluation. Each of these codes are implemented in pytorch framework. We evaluate each one of them on a single NVIDIA A100 40GB GPU. For closed source Video-LMMs, we use their respective API for evaluation. We use GPT-4o~\cite{gpt4-o} as our LMM to generate the captions and the final QA pairs in VANE-Bench. Next, we describe the prompts used in our Caption Generation Module (CGM), Question Answer Generation Module (QAGM), and in evaluating various Video-LMMs on VANE-Bench in the subsequent subsections.

\subsection{Caption Generation Module (CGM)}
\label{sec:cgm-prompt}
\textbf{System Prompt:} You are a helpful and intelligent AI assistant which can generate informative captions for a given input of 10 consecutive images/frames from a video. The video is generated from an AI text-to-video diffusion model and has some obvious inconsistencies or anomalies in the form of various deformations, unrealistic physical transformations, unnatural appearance of objects, human faces, body parts, or sudden appearance, disappearance, or merging of objects. Your task is to generate a descriptive caption for the given input video, highlighting the inconsistencies or anomalies present in the video.

\noindent
\textbf{Text Prompt:} Please generate a detailed caption which describes all the given frames. Some of the frames may contain inconsistencies which are annotated with a green bounding box around them with the type/name of the inconsistency. Your generated caption should capture the details of the entire video, while also describing all the inconsistencies. Thus, properly look at all the given frames and the region marked by the green bounding boxes when describing the inconsistencies. Further, make sure to mention specific details about each of the inconsistencies, and mention the exact names of the inconsistencies from the marked green bounding box. Also, while describing the inconsistency please be as specific and detailed as possible, don't be vague or general about the inconsistency. The reader of the caption should perfectly understand what inconsistencies/anomalies are in the video and what the video is about. Do not mention the green bounding box in your response; it is only for you to identify the inconsistencies. Make sure to describe all the inconsistencies in your caption. Do not analyze the impact of the inconsistencies; you should only describe them. There is no need to mention when the inconsistencies start or end, just describe them.

\subsection{Question Answer Generation Module (QAGM)}
\label{sec:qagm-prompt}
\textbf{System Prompt:} You are a helpful and intelligent AI assistant which can curate high-quality and challenging question and their corresponding answers, which are used to test the video understanding capabilities of an multi-modal LLM model capable of taking videos as their inputs.

\noindent
\textbf{Text Prompt:} You are given a video input, which is generated by a state-of-the-art AI algorithm. Thus, these videos look very natural and almost realistic, but they are actually synthetic and generated by an AI algorithm. The videos may have some inconsistencies or anomalies present in them, which are generally localized to only a specific location in the video as identified by the green bounding boxes in the video. The rest of the video appears completely natural or realistic. This specific inconsistency may last for only a few frames of the video or may last for the entire video itself. The inconsistency or anomalies in the video are generally events and phenomena which is not observed in real-world and physical scenarios. You will also be given a caption as input that describes the video, along with the specific inconsistency present in the video. Based on the given video and caption input, your task is to formulate 3 diverse and misleading questions to test whether the multi-modal LLM model can correctly identify the options based on the inconsistencies present in the video or not. So, your generated questions should give the model few options to choose from to make its answer, and these options should be of high quality and also have misleading choices so that you can test deeper level of understanding of these multi-modal LLM models. Thus, the goal of these questions is to accurately assess the multi-modal LLM's ability to accurately identify the inconsistencies present in the video. Generate questions that comprise both interrogative and declarative sentences, utilizing different language styles, and provide an explanation for each. Your response should be presented as a list of dictionary strings with keys 'Q' for questions and 'A' for the answer. Follow these rules while generating question and answers:

1. Do not provide answers in the question itself. For example, the ground-truth attribute or component that makes the video scene unusual should never be mentioned in the question itself.

2. Ensure the questions are concrete and specific, and not vague or ambiguous.

3. The questions should be formed based on your deep understanding of the video and the caption. Thus, properly read the caption and look at the given video to generate the questions.

4. The questions should only pertain to the inconsistencies present in the video, and not about the video in general.

5. You may also ask the model some misleading questions talking about non-existent inconsistencies in the video, to test the model's ability to differentiate between real and fake inconsistencies.

6. Do not ask vague questions, and the answer should only contain one of the correct option mentioned in the question.

7. In your question itself you must provide multiple choice options for the answer, and the answer should be one of the options provided in the question. Please ensure you provide option choices and their corresponding letters in the question itself.

8. In your answer, only mention the correct option letter from the question. Make sure that the correct option letter is not always the same, and randomly shuffle the correct option letter for each question.

9. You must only follow the below output format and strictly must not output any other extra information or text. Your output format should be strictly as follows, without any additional information or text:

[{"Q": 'first question A) <option1> B) <option2> C) <option3> D) <option4>', "A": 'Pick the correct option letter from A) B) C) D)'}, {"Q": 'second question A) <option1> B) <option2> C) <option3> D) <option4>', "A": 'Pick the correct option letter from A) B) C) D)'}, ... \}]

Given below is the caption input which describes the given video along with the specific inconsistency present in the video. The caption is: \{caption\}

\subsection{Evaluating Video-LMMs}
\textbf{System Prompt:} You are a helpful and intelligent multi-modal AI assistant, capable of performing visual question-answering (VQA) tasks. You will be given as input 10 consecutive frames from a video, and a corresponding question related to the video, you have to answer the given question after analyzing and understanding the given input video. The question itself will present you with 4 lettered options like A) B) C) D), your task is to only output single letter corresponding to the correct answer (i.e. string literal 'A', 'B', 'C', or 'D'), and you should not output anything else.

\noindent
\textbf{Text Prompt:} \{question\}

\section{Distribution of VANE-Bench dataset}
\label{appendix:vane-bench-distribution}
\textbf{How to view the dataset?} The dataset alongside metadata will be hosted on the Hugging Face platform for download post acceptance of the paper. Users can directly load the dataset using Hugging Face Datasets library or download the zip file in the same Hugging Face repository. All instructions and code files to reproduce the experiments of the paper will be provided in a github repository.

\noindent
\textbf{How will the dataset be distributed?}  
The dataset will be distributed to the public using the Hugging Face Dataset Hub. We have publicly released the codebase alongside instructions to reproduce and evaluate models on GitHub. 

\noindent
\textbf{Dataset License.} This work and dataset is licensed under a Creative Commons Attribution-NonCommercial-ShareAlike 4.0 International License. The videos in the VANE-Bench dataset are collected from publicly available sources and existing real-world datasets and are for academic research use only. The video generative models used to synthesize data samples in our VANE-Bench benchmark are open to use publicly and do not pose any privacy concerns as the persons or objects present in the generated videos are synthetic and do not exist in the real world. The real-world surveillance datasets - UCFCrime \cite{ucf-crime}, UCSD Pedestrian \cite{ucsd-ped}, Avenue \cite{cuhk-avenue}; on the other hand, used in our work are all existing well-known and publicly available datasets that are released under open-source licenses. Thus, the original creators of these datasets have collected the data after taking informed consent from the stakeholders. By using VANE-Bench, you agree not to use the dataset for any harm or unfair discrimination. Please note that the data in this dataset may be subject to other agreements. Video copyrights belong to the original dataset providers, video creators, or platforms.
\end{document}